\documentclass[sigconf,authorversion]{acmart}
\usepackage{cleveref}
\usepackage[inline]{enumitem}
\usepackage{bm}
\usepackage{subcaption}
\AtBeginDocument{%
  }

\newcommand{\ie}{\textit{i.e.}}
\newcommand{\eg}{\textit{e.g.}}

\renewcommand{\S}{\mathcal{S}} 
\newcommand{\A}{\mathcal{A}} 
\newcommand{\RW}{\mathcal{R}} 
\newcommand{\T}{\mathcal{T}} 
\newcommand{\R}{\mathbb{R}} 

\copyrightyear{2023}
\acmYear{2023}
\setcopyright{acmlicensed}
\acmConference[KDD '23]{Proceedings of the 29th
ACM SIGKDD Conference on Knowledge Discovery and Data Mining}{August
6--10, 2023}{Long Beach, CA, USA}
\acmBooktitle{Proceedings of the 29th ACM SIGKDD Conference on Knowledge
Discovery and Data Mining (KDD '23), August 6--10, 2023, Long Beach, CA,
USA}
\acmPrice{15.00}
\acmDOI{10.1145/3580305.3599777}
\acmISBN{979-8-4007-0103-0/23/08}

\acmSubmissionID{403}

\begin{document}

\title{Automatic Music Playlist Generation via Simulation-based Reinforcement Learning}

\author{Federico Tomasi}
\email{federicot@spotify.com}
\affiliation{%
  \institution{Spotify}
  \city{London}
  \country{United Kingdom}
}

\author{Joseph Cauteruccio}
\email{jcauteruccio@spotify.com}
\affiliation{%
  \institution{Spotify}
  \city{Boston}
  \country{USA}
}

\author{Surya Kanoria}
\email{suryak@spotify.com}
\affiliation{%
  \institution{Spotify}
  \city{San Francisco}
  \country{USA}
}

\author{Kamil Ciosek}
\email{kamilc@spotify.com}
\affiliation{%
  \institution{Spotify}
  \city{London}
  \country{United Kingdom}
}

\author{Matteo Rinaldi}
\email{matteor@spotify.com}
\affiliation{%
  \institution{Spotify}
  \city{New York}
  \country{USA}
}

\author{Zhenwen Dai}
\email{zhenwend@spotify.com}
\affiliation{%
  \institution{Spotify}
  \city{London}
  \country{United Kingdom}
}








\renewcommand{\shortauthors}{Federico Tomasi et al.}

\begin{abstract}
Personalization of playlists is a common feature in music streaming services, but conventional techniques, such as collaborative filtering, rely on explicit assumptions regarding content quality to learn how to make recommendations. Such assumptions often result in misalignment between offline model objectives and online user satisfaction metrics. In this paper, we present a reinforcement learning framework that solves for such limitations by directly optimizing for user satisfaction metrics via the use of a simulated playlist-generation environment. Using this simulator we develop and train a modified Deep Q-Network, the action head DQN (AH-DQN), in a manner that addresses the challenges imposed by the large state and action space of our RL formulation. The resulting policy is capable of making recommendations from large and dynamic sets of candidate items with the expectation of maximizing consumption metrics. We analyze and evaluate agents offline via simulations that use environment models trained on both public and proprietary streaming datasets. We show how these agents lead to better user-satisfaction metrics compared to baseline methods during online A/B tests. Finally, we demonstrate that performance assessments produced from our simulator are strongly correlated with observed online metric results.
\end{abstract}



\maketitle

\section{Introduction}
Generating personalized playlists programmatically enables a dynamic form of music consumption that users expect from major music streaming platforms. Machine learning (ML) methods~\cite{Bonnin2014-jk} are commonly used to power personalized playlist experiences that attempt to optimize for both content quality and users' musical preferences.
Music playlist personalization and generation methods often rely on methods such as collaborative filtering~\cite{Vall2015-yd, Nabizadeh2016-zb, Kaya2018-rb} or sequence modeling~\cite{Choi2016-vd, Irene2019-ji}.
These methods make strong content quality assumptions that can lead to various practical limitations.
For example, explicit feedback based collaborative filtering assumes that a "good" playlist consists of tracks to which a user would assign a high rating and, as a result, often struggles to consider other important factors such as acoustic coherence, the context of a listening session, and the potential presence of optimal item sequences.
Conversely, in industrial music streaming systems, quality is assessed through metrics derived from user activity, such as average streaming time or the number of days during which user was active during a given period.
The limitations arising from modeling assumptions in conventional methods can lead to a mismatch between offline metrics and user satisfaction metrics.
For example, a collaborative filtering method may return a playlist with the highest predicted ratings but contains a mixture of adult and kids music, which usually does not lead to good user satisfaction.
This makes developing a good playlist generation system challenging in practice.

Reinforcement learning (RL) offers an orthogonal approach that does not require explicit quality assumptions but can instead interact with users to learn a playlist generation model that directly optimizes for satisfaction.
RL agents interact with users to explore and identify important factors that drive playlist quality as assessed by consumption metrics and, as such, can overcome the target-to-metric disconnect of conventional playlist personalization methods.

In order to apply RL to music playlist generation, the generation problem needs to be formulated as a Markov decision process (MDP), in which states encode contextual information summarizing a user listening session, the action space is the space of all the possible playlists and the reward is the desired user satisfaction metric. 
A major challenge of this formulation is the combinatorially complex action space. 
For example, the task of generating a playlist of 30 tracks drawn from a pool of 1000 candidates results in a discrete action space with about $10^{89}$ possibilities, 
which is significantly larger than the action spaces of common game-focused RL problems such as GO and Atari.
This means that off-the-shelf RL methods for discrete actions are not readily applicable to the playlist generation problem. 
In related recommender system problems, such as slate recommendation, slate-MDP~\cite{Sunehag2015-qb} and slateQ~\cite{Ie2019-po} have been proposed to tackle this combinatorial action space challenge.
In slate recommendation, a list (or slate) of items is recommended to a user after which the user picks the single best item from the slate.
Both methods~\cite{Sunehag2015-qb,Ie2019-po} rely on the assumption that a user selects one or zero items from the slate to decompose the MDP for efficient learning. This critical assumption is violated in our problem space since we desire users to consume a significant portion of the presented playlist (slate). 
The resulting inability to decompose the MDP makes the correct application of methods proposed by \cite{Sunehag2015-qb,Ie2019-po} impossible in our setting.

\paragraph{Contributions.}
In this paper, we propose a simulation-based RL approach for music playlist generation. 
We first propose the use of a user model. This model is trained using user listening sessions and is used as part of a simulated environment to estimate the outcome of user interactions with playlists generated by an RL agent.
In the simulated environment, the agent performs a generation task by consecutively choosing single tracks to recommend to the user. In turn, we use the user model to predict how a real user would respond to that track, and pass the information to the agent prior to choosing the next track to add to the sequence.
The action space in this simplified setting is the size of track candidate pool. 
The iterative selection of single tracks from the candidate pool makes the agent training tractable.

The crucial component of our simulator design is the user behavior model, or user model. This model distills user behavioral patterns from complex listening data allowing for the creation of a simplified but realistic simulation environment for agent training. 
This enables us to experiment with RL agents in a simulated environment and accurately assess their performance prior to online experiments where real users are exposed to the resulting policies.

Using such a user model, we develop a modified Deep Q-Network (DQN) agent for music playlist generation. 
Differently from standard DQN agent formulation, 
the agent ingests item (track) features and returns the associated Q value,
thus enabling Q-value estimates for items unseen during training. 
This formulation allows us to train a single DQN to generate playlists from a diverse set of candidate pools consisting of hundreds of millions of tracks overall.  
The performance of the this agent was evaluated offline using proprietary Spotify systems in addition to a publicly available Spotify listening sessions dataset. 

The developed agent was also evaluated in online A/B tests. With these A/B test results, we then show that the policy performance evaluations from our simulated environment strongly correlate with online user satisfaction metrics, thus validating our approach. 
The main contributions of this paper are summarized as follows:
\begin{enumerate*}[label=\emph{(\roman*)}]
\item We propose a new RL approach for music playlist generation based on user simulation, which allows us to decompose the combinatorial action space to a sequence of tractable actions.
\item We develop a user behavioral model based on a recurrent neural network that learns sequential dynamics of user interactions during music listening resulting in better accuracy than non-sequential user models.
\item We develop a modified DQN agent whose policy, when trained in a simulated environment for playlist generation, shows good offline and online performance.
\item Using online tests we show that our agent leads to better user-satisfaction metrics than baseline methods and that performance assessments from our simulator are strongly correlated with these online metric results.
\end{enumerate*}


\section{Related Work}\label{sec:related}
Music recommendation is a topic of increased interest in recent years \cite{singh2021neural,chang2021music}. Recommending music on a streaming platform is a non-trivial challenge. It is especially difficult when compared to other types of content
because implicit music preferences of users are hard to model in a generalized fashion \cite{kordumova2010personalized}.
Moreover, music streaming users are more likely to listen to the same songs several times (in comparison to buying the same items on e-commerce platform or watching the same movies on video streaming platforms being relative less likely), so the trade-off between exploration and exploitation (\ie, recommending new tracks or recommending tracks the users are familiar with) becomes harder to balance. 
RL approaches provide theoretically valid and practically proven mechanisms to solve such problems and, as such, have been extensively explored for music recommendation tasks \cite{wang2014exploration,wang2020hybrid,liebman2014dj,hu2017playlist, Sakurai2022-qm, Sakurai2020-dp, Shih2018-kz}.
Prior work aims at modeling user preferences as part of the RL procedure itself, however; there is no generally accepted procedure used to translate user musical preferences into an actionable reward for agents. Some approaches require that users score each song in the dataset (\eg, \cite{hu2017playlist}), which would be infeasible in large scale applications. Others relax this requirement and divide songs in the catalog into predefined bins before assigning each user to one or multiple clusters (\eg, \cite{liebman2014dj}), which, conversely, may not be fine-grained enough to capture the nuances of the user preferences. Traditional methods that use a static notion of preference are also limiting as user preferences are inherently contextual, so the same types of songs may be relevant in particular situations but not in others (\eg, sleep music while exercising).
Current work that attempts to use RL methods for music recommendation also fails to sufficiently account for the sparsity of user-item signals in lean-back interaction modalities. Moreover, due to the low-friction item interaction cost of music to users, it is challenging to use implicit or explicit signals to derive a generally valid content rating \cite{kordumova2010personalized}.

This work combines an accurate model of the environment (a critical component of our offline RL formulation) with RL agents best suited for non-myopic decisions. Our model provides us with a generalizable mechanism via which we can approximate complex user behaviors and translate them into a reward function that the agent can effectively use to learn a satisfactory policy. 
This, coupled with our agent design, allows us to generate playlists customized to different types of users in a manner that maximizes their expected satisfaction as determined by our metrics. 

Previous RL for music playlist generation methods~\cite{liebman2014dj,hu2017playlist, Sakurai2022-qm, Sakurai2020-dp, Shih2018-kz} consider the action space to be a single track, which is similar to the MDP of our simulated environment. 
The major difference is that these methods learn an agent directly from recorded user listening sessions.
This training paradigm implies that a user goes through the exact list of items chosen by the agent in the presented order.
In reality, a user interaction with a playlist is much more complex, \eg, a user may listen to the generated playlist with a random shuffle or manually pick individual songs to listen to.
We propose instead training a user behavior model from recorded user listening sessions and use it to build a simulated environment.
Then, an agent is trained in the simulated environment.
This approach avoids the issue of MDP assumption violation in the previous methods.

In particular, this work is based on two active branches of research: model-based RL \citep{sutton2018reinforcement} and recommender systems \citep{jannach2010recommender, aggarwal2016recommender}. 
Our approach to model-based RL is loosely based on the Dyna architecture \citep{sutton1991dyna}. 
The Dyna architecture uses data gathered from real user interactions to learn an environment model that is in turn used to improve the policy. 
However, unlike \citep{sutton1991dyna}, we decouple the step of pre-training the user model from training the agent using that model. 
This is done mainly to circumvent engineering constraints but has the additional benefit of a more parsimonious architecture. 
The fact that we use a neural network to learn the user model also makes our system similar to other methods that use deep learning to predict dynamics \citep{werbos1989neural, kumpati1990identification, wahlstrom2015pixels,oh2015action, moerland2020model}.  

The recommender systems community has increasingly embraced RL \citep{afsar2021reinforcement} as a way of generating ranking policies which are good quality and less myopic than more traditional solutions such as collaborative filtering methods and learning-to-rank. Our work is similar to the pioneering approach by \citet{sunehag2015deep} in that we use deep learning to approximate $Q$-functions, but unlike \citet{sunehag2015deep}, we do not require the slate-MDP formalism.
Our approach can be thought as evaluator-based, aiming at estimating the optimal action-value function through the use of a deep Q-network.
While few prior works combine Q-learning and an environment model into a recommender system (\eg, \citep{afsar2021reinforcement}),
the exact specifics of our application, mainly having a user-generated signal for each component of the slate coupled with the scale at which our model was deployed, make our work distinct.

\section{Problem Formulation}\label{sec:problem-formulation}

Consider the problem of automatic playlist generation in a music streaming platform. 
The catalog is composed of millions of songs (often called \textit{tracks}) created by a diversity of artists. 
Each track and artist is associated with a variety of features that summarize nuances useful when recommending or otherwise modeling content (traits intrinsic to the songs themselves, such as audio features like beats-per-minute or key, or aggregate user consumption traits). 
In this work, we assume that we have access to such features for each recommendable track. 
Further details on how to generate such features can be found in previous works (\eg, \cite{brost2019music}).
Here we consider the automatic playlist generation problem, where users decide on a general super-set of content they are interested in (such as a genre, \eg, ``indie pop''), and the platform is tasked with generating a playlist that satisfies users given this initial interest.
Having a catalog of millions of tracks, it is unlikely that such a playlist already exists especially for tail content types. 
Further, each user experiences music differently, hence automatic playlist generation is a relevant and practical problem for music streaming platforms to create the best personalized experience for each user.

In order to design a method to best pick a list of tracks for the user, we first design a user behavior model that estimates how a user would respond to these tracks. Using this model we can optimize the selection of tracks in such a way as to maximize a (simulated) user satisfaction metric.
To model users, we assume that each user has an implicit prior preference for tracks belonging to certain categories. For example, one user may usually like \textit{rock} music, while another user may prefer \textit{hip hop}. However, we do not assume that (static) genre preference is the only factor that influences which tracks a user wants to listen to. 
Instead, we consider multiple other factors, such as contextual preference based on time of day, device type and so on (for example, music patterns during commute hours may differ than music patterns before sleep, or during working hours) \cite{hansen2020contextual}.

Further, we consider a subset of tracks that are available for each playlist based on the general preference of the user, which we refer to as the \textit{candidate pool}. For example, based on the genre ``indie pop'' we utilize a pool of candidates that have been previously assigned to this genre. Each candidate pool has a dimensionality much lower that the size of the whole catalog of available tracks in the platform.
Utilizing these components we solve the recommender problem using a model-based RL framework that we detail in the next section.

\section{Methodology}\label{sec:methodology}
\begin{figure}[t]
    \centering
    \includegraphics[width=0.9\linewidth]{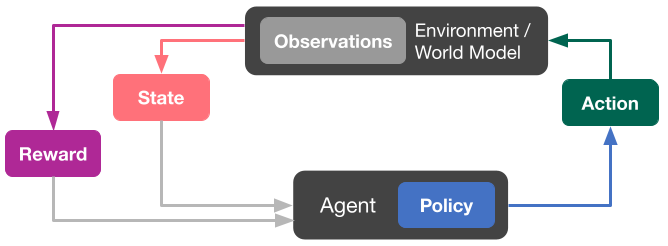}
    \caption{Reinforcement learning loop in a simulation-based environment.}
    \label{fig:rl-loop}
\end{figure}

\Cref{fig:rl-loop} illustrates the main training loop in a model-based RL framework \cite{moerland2023model}. 
The environment consists of a \textit{world model} that models the user behavior in response to an action $a_t \in \A$ at timestep $t$, which is also referred to as a user behavior model in the previous text.
This model captures the state transitions given the action selected from the agent and the state information visible to the agent.

More precisely, the environment models a transition function $\T: \S\times\A\to \S$ and a reward function $\RW:\S\times\A\to \R$. 
The agent is tasked with learning a policy that selects actions to perform on the environment based on the current state of the environment. The policy is hence represented as a function $\pi: \S \to \A$, that maps a states to actions. 
The first state is sampled from the initial state distribution $p(s_0)$ by the environment and observed to the agent. 
After the action is selected by the policy, the environment returns the next state $s_{t+1} \sim \T(\cdot | s_t, a_t)$.

We consider a recommendation task, in particular for content programming in a music streaming domain. In such framework, the action is the item (\ie, the song) recommended. The environment consumes the action proposed by the agent and uses the world model to transition to the new state and returns a specific reward conditioned on the action (for example, whether the user adds the song to their favorites or plays the item).
The agent is able to see the new state $s_{t+1}$ and reward $\RW(a_t)$, using them to adapt the policy and predict the next action to pass to the environment for the next iteration. 
This procedure continues until the world model signals the termination of a specific sequence of states (\textit{episode}).

During the training phase the environment makes use of a \textit{user model} to return a predicted user response for the action recommended by the agent. The user model takes the action (track) and its features as inputs and makes a prediction based on the current state of the environment. 
Such user models are designed to distill user behavior from real (past) user interactions with the platform. 
Further, the state includes information such as user specific features and the history of the actions already selected by the agent.

In this setting, the effectiveness of policy that results hinges on the accuracy of the offline environment. Making real word recommendations using the agent policy requires the user model to accurately reflect real-world user preferences. 
In what follows, we expand on how we design the world model for this task.

\subsection{World Model Design}\label{sec:world-model}
To accurately learn the policy in an offline setting, the agent needs to interact with a model that mimics a real recommendation scenarios 
as close as possible to the same conditions the agent will encounter when deployed online.
To accomplish this, we design a \textit{world model} that includes the information of the candidate pools, the user state information, as well as the current track features. 
The world model (environment) models a transition function using historical data, and uses this function, once trained, to produce a new state from the action selected by the agent as well as the reward that results from the specific action.

The world model is responsible for starting the session (start of episode), keeping track of list of tracks recommended to the user, for terminating the session (end of episode). The combination of all of such components is how the world model simulates a user session.
Further, in order to model the reward, the world model utilizes a \textit{user model} to predict the user response to an item. The user model is designed to be as accurate as possible in its ability to predict how users would have behaved in response to the actions selected by the agent, by distilling user preference.
Such user model is a supervised classifier trained using data collected from real users during past online experiments.
We use a randomized policy to collect ground truth interaction data for training. Specifically, we randomly shuffle the songs in a number of music listening sessions and collect outcome information for each track streamed by the user, such as the percentage of the track completed by the user (if streamed) and the presentation position of the track within the listening session.

To these data we join user information (\ie, features related to user interest and past interaction with the platform), which we call \textit{context features}, and content information (\ie, information related to the content that the user is interacting with), which we call \textit{item features}. The model uses these features to predict a set of \textit{user responses} for each track. 
In practice, we perform this task by training a classifier with multiple outputs, each one mapped to a response. The model is optimized to predict three different (related) user responses: 
\begin{enumerate*}[label=\emph{(\roman*)}]
    \item whether the user will complete the candidate track, 
    \item whether the user will skip the candidate track, 
    \item and whether the user will listen to the track for more than a specific number of seconds (specified a priori).
\end{enumerate*}
The actual implementation of the user model changes the complexity and the modeling power to mimic real users.
While different parameterizations of the user model are possible, 
we design two different user models, a sequential and non-sequential model, which we refer to as SWM and CWM, respectively. 

The sequential model is a sequence of LSTM cells\footnote{Different parametrizations can be applied to capture the sequential information, such as transformers \cite{vaswani2017attention}. However, detailed comparisons about different world model implementations are outside of the scope of the paper.} that capture the order of tracks within the same episode and use the user response at track $t-1$ to predict the user response at track $t$. 
More specifically, the sequential user model defines a (auto-regressive) sequential model of the form:
\begin{align}
    p(\bm y_{(0, \dots, T)}) = \prod_t p(y_t | y_0, \dots, y_{t-1}, i_0, \dots, i_{t}, u),
\end{align}
where $i_t$ represents the item features of the track at the $t$-th position, $y_t$ represents the user response that corresponds to track $i_t$ and $u$ represents the context information (user information in addition to other information about the current session). 
The parameterization of $p(y_t)$ uses a LSTM to capture the sequential information from $0$ to $t-1$. Empirically, we consider a sequential 3-layer model with (500, 200, 200) LSTM units.

For the non-sequential user model, we construct a series of dense layers where the information in the input (features) is limited to features that summarize the track and the user with no other information on the session itself (such as position of the track within the session, or user responses of previous tracks in the session) provided to the model. The non-sequential user model takes the following form:
\begin{align}
    p(\bm y_{(0, \dots, T)}) = \prod_t p(y_t | i_t, u),
\end{align}
where response $y_t$ to track $i_t$ does not depend on previous tracks, 
and the probability decomposes into independent components.

Each model has its own advantages. The non-sequential model is simple and fast which makes it easier to use for agent training. The sequential model is more accurate in the classification task due to its access to sequential information coupled with the fact that user responses to tracks in sequence are often correlated.
Empirically, we found that the use of a sequential model (with intrinsically higher stochasticity) makes training the agent far more difficult and thus can be less beneficial in practice.
Additionally, the non-sequential model allows us to easily compute the maximum theoretical reward since it is not dependent on the actual order of songs shown to the users. For this reason, during offline evaluation we are able to understand how far the agent is from the maximum expected reward.

At each action picked by the agent, then, the world model collects the information about the specific track (track features) as well as context information (user response history, features of previous tracks in the session and so on) to pass to the user model and compute the predicted user response. The user response is then translated into the reward that is passed to the agent. 

We also note that the user models can be used directly for item selection thus effectively bypassing the requirement of learning a policy using the RL agent. Direct predictions from world models such as the CWM serve as a strong baseline for our agent policies. To use these user models for inference we use a planning policy that turns user-track scores into a sequence of tracks. 
This is done using a greedy ranking algorithm, the application of which to a model is typically referred to as greedy model predictive control (GMPC). 
When combined with the non-sequential user model CWM, we refer to such planning policy as CWM-GMPC in our experiments.
Note that the optimal policy when using the CWM for agent training is equivalent to CWM-GMPC, since item order does not affect predictions. As such, our expectation is that online, agents trained using CWM in simulation should be statistically indistinguishable from the CWM-GMPC baseline. 

\subsection{Action Head DQN Agent}\label{sec:agent}
A highly diverse set of musical interests across users and user groups coupled with the potential for varied and sometimes even orthogonal listening behavior or objectives (\eg, consuming familiar or unfamiliar content) on the part of individual users requires a recommender system that can both generalize and adapt. RL methods meet these requirements and we propose a simple and efficient RL solution based on Deep Q- Learning (DQN) \cite{mnih2013playing}. 

The DQN agent uses a deep neural network to predict the recommendation quality (Q) of each item (action) in a listening session. The main idea behind the Q network is that each available track is assigned to a specific value (a \textit{Q value}), and the track with the highest value is then selected by the agent. The reward as returned by the environment after each action is used to update the Q network.

An intrinsic limitation in our use case is that the pool of candidate tracks is dynamic and may depend on contextual or user information.
In particular, the agent needs to be able to generalize its selection strategy across different candidate pools. This requirement results in our DQN differing from typical formulations where the Q network, given a state representation as input, produces an array of quality predictions where each array element maps to a specific action. This is not applicable in our case since the item space as defined by the pool is constantly changing, and therefore, so is the output space of the Q network and its mapping to specific actions.
Simultaneously, we need to train a single agent that can operate across a diverse set of candidate pools that in practice can have little to no overlap. Empirically choosing to train a single agent makes more efficient use of our training data and generated episodes.
Practically, it also means that in production we need to maintain only one agent and not one agent per pool, user, or pool-user combination. 

\begin{figure}
    \centering
    \includegraphics[width=0.9\linewidth]{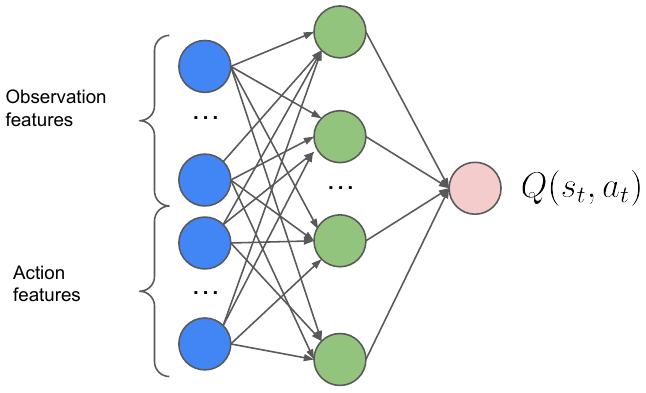}
    \caption{Action-head network. Both observation and action features are taken in input in order to estimate the Q value.}
    \label{fig:q-net}
\end{figure}
Because of such constraints we developed what we refer to as an \textit{action head} (AH) DQN agent (\Cref{fig:q-net}). 
Our AH-DQN's Q network takes as input the current state and the list of feasible actions. The network will produce a single Q value for each action, and the one with the highest Q value is selected as the next action to apply.
This ensures that a single Q network can be optimized independently from the set of available items to recommend during an episode.
We model the different environments as a contextual MDP \cite{hallak2015contextual}, to ensure generalization across users and candidate pools thus allowing for the existence of one single agent. 
Intuitively, this approach generalizes well to arbitrary and even dynamic candidate pools. The downside is that we need to make a forward pass through the Q network for each allowed action in the candidate pool to predict its Q value. However, as the candidate pool is much smaller than the catalog of items to recommend, this is still efficient to do in practice. Furthermore, this step could be efficiently parallelized as the forward passes through the Q network, for each action at a step, are independent from each other.

Our goal is to optimize the following policy: $\pi_{\text{AH}}: \S \times \A^L \to \A$, which takes as input the state \textit{and} the list of available actions, denoted $\A^L$, and selects the best action according to the computed Q values:
$$\pi_{\text{AH}}(s_t) = \arg\max_a Q(s_t, a).$$

The Q network makes use of the Bellman optimality equation to estimate the quality of an action at time $t$, $a_t$ as follows:
\begin{align}
Q(s_t, a_t)=r(s_t, a_t)+\gamma \max _a Q\left(s_{t+1}, a\right)
\end{align}
where $s_t$ is the current state, $r(s_t, a_t)$ is the reward estimate, and $s_{t+1}$ is the next state.

The state of a user $s_t \in \S$ encodes all the information available to the agent at time $t$. $\S$ is the set of possible states. The state variable $s_t$ obeys the Markov property $p(s_{t+1} | s_t, a_t) = p(s_{t+1} | s_1, a_1, \dots, s_t, a_t)$, since $s_t$ encodes all the information in $(s_1, a_1, \dots, s_{t-1}, a_{t-1})$.

\paragraph{Reward and User Simulator.}
Our reward function $\RW: \A \to \R$ associates the recommended track at step $t$ with a measure on how successful the recommended track (or sequence of tracks up to and including time-$t$) is with respect to some user outcome. In our experiments, we measure the performance of the recommender system based on the probability of a track to be completed by the user, so $\RW(a_t) \in [0, 1]$.
In our model-based RL framework, the user behavior is approximated using a user model (as described in \label{sec:world-model} above) and, in this setting, the probability of completion estimated by the model serves as a proxy for the real user preference.
Then, for each episode we compute the sum of completion probabilities as predicted by the user model until episode termination, which is the total reward for the episode. The agent updates the Q network at each iteration to maximize this reward.

\section{Experiments}\label{sec:experiments}
We tested our model on both public and proprietary streaming datasets. We evaluate public streaming performance in an offline setting only. For our online experiments, we train a user model on proprietary streaming data of similar structure to the public dataset, train the agent as described, and deploy a fixed agent policy in an online A/B testing setting.

\subsection{Public Streaming Dataset}\label{sec:public-results}
We first use a public music dataset from Spotify \cite{brost2019music} including 160 million music listening sessions to empirically evaluate our proposed model-based RL formulation. 
Features that describe tracks (such as acoustic properties and popularity estimates) and sessions are provided in the dataset.
We organize our data by ordering session interactions over time. At each interaction there are indicators in the dataset that note if the track was completed or skipped at or before three track-time markers: $r_1, r_2, r_3$. Using these data we design simulation environment to train and test our agent. First we train a response model that takes as input a sequence of tracks up to time $t-1$, $\{a_i, i=0, \dots, t-1\}$ and an action-track at $a_t$. It then predicts the probability of a skip outcomes $r_1, r_2, r_3$ for $a_t$. 

We regard each listening session as a list of tracks from a unique playlist that the agent aims at creating. As described in \Cref{sec:problem-formulation}, a user (associated to its listening session) will initiate a play session by selecting a musical sub-context (such as a genre). Each sub-context has a pool of associated items (tracks) from which we draw candidates to generate a playlist. The user will then interact with the playlist until the selected tracks are exhausted (one source of episode termination)\footnote{Note that, in online scenarios, the user may quit the session prior to consuming all candidates in the playlist. For simplicity we use a fixed size termination criterion during offline experiments and evaluation, but other termination criteria remain possible.}.

We impose some design specifics for our simulation. First, we sample sessions of length 20 from the data. At the first step the environment observation is composed of the first 5 tracks in the play session in the order observed. At each step the agent action is to pick one track from the remaining 15: the response model then provides a predicted outcome for this selection (probability of completion). The agent is rewarded for picking tracks with a low likelihood of being skipped as determined by the response model and the episode is terminated after 15 steps. As such the maximum achievable reward in our simulations is the maximum of the sum of completion probabilities for each session, which is upper-bounded by the size of candidates in the playlist, \ie, 15.

We use a non-sequential user model (CWM), to simulate user responses during agent training. 
As no user features are available in the public dataset (users are anonymized for privacy), 
we utilize only on the track features and past user responses in the session to compose our feature space. 
Specifically, the user model utilizes first 5 tracks in the play session, represented by their features, 
as a context feature to estimate the outcome of the remaining 15 tracks, each represented by their respective features. 

We then train the agent as introduced in \Cref{sec:agent} against the CWM. 
As mentioned in \Cref{sec:world-model} this allows for more tractable and faster training than more complex user models. 
We train the Q-network using an element-wise mean squared error TD loss. Such loss is implemented using an additional target network with the same structure of the original Q network and its own parameters and layers, which is updated with a period of 50 that empirically stabilizes the training.
We compare our method to two baselines:
\begin{enumerate*}[label=\emph{(\roman*)}]
    \item \emph{Random}: a model to randomly sort the tracks by shuffling the 15 remaining tracks in the session with equal probability; and 
    \item \emph{CWM-GMPC}: a combination of the non-sequential user model and a greedy ranking policy, as described in \Cref{sec:world-model}
\end{enumerate*}
Finally, we use an independent metric model to estimate the policy performance of the agent against the random and CWM-GMPC baselines.
To better simulate user behavior during offline evaluation, we use the sequential world model (\textit{SWM}) as introduced in \Cref{sec:world-model}. 

\begin{table}[]
    \centering
    \caption{Average return for random and agent policy on offline evaluation of the public streaming dataset.}
    \label{tab:public-streaming}
\begin{tabular}{lccr}
\toprule
             Policy &  Avg. Return &  $\sigma$ &       CI (95\%) \\
\midrule
 Action Head Policy &         1.94 &   1.27 &  (0.32 | 4.18) \\
           CWM-GMPC &         2.46 &   1.10 &  (0.83 | 4.36) \\
             Random &         0.98 &   0.76 &   (0.0 | 2.54) \\
\bottomrule
\end{tabular}
\end{table}
\begin{figure}
    \centering
    \includegraphics[width=\linewidth]{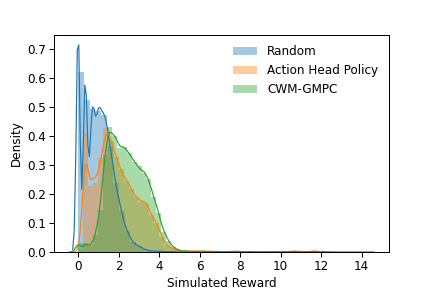}
    \caption{Reward distributions for the compared policies on the public streaming dataset as estimated by the SWM.}
    \label{fig:pubdist}
\end{figure}

\Cref{tab:public-streaming} includes the result summary of our comparisons between our agent and the two baselines using SWM as the evaluation procedure.
Note that each policy results in relatively modest simulated average returns (between 1 and 2.5). This is because of two reasons. First, the dataset does not contain any user preference features and any non-sequential model has access to only track features during inference. Second, the action space is made up of only 15 predefined tracks and the goal is to sort them accurately. Our trained policies are non-sequential, their ability to achieve highly accurate results on such a task will be somewhat limited. We note that even the CWM's greedy optimal ordering produces better predicted completion counts form the SWM than a random policy. This is clearly demonstrated by the fact that the CWM-GMPC has the highest average return among the three methods tested. CWM-GMPC orders tracks optimally according to the simulated responses of CWM. Since the agent is trained against the CWM in simulation this provides a bound on the performance of the agent.

This bound manifests in the agent. The policy is able to outperform a random shuffle, implying that the agent policy is has captured some information while training in simulation with the CWM that enables it to provide a better track ordering than random.

We also plot the distribution of rewards in \Cref{fig:pubdist}. The distributions show how a random policy mostly fails to return a satisfactory list of tracks, further validating the assumption that the order of tracks is indeed meaningful for a good listening experience. The policy learned by our proposed agent is able to provide a better experience than the random policy for the majority of the users, concentrating the majority of (simulated) rewards between 2 and 4 (which correspond on average to at most 7 and 9 tracks completed in the session of length 20).
We can notice a spike in position 0 of simulated rewards for the agent policy which is not shared from the CWM-GMPC. 
This highlights that, for a few users and sessions, 
the agent fails to predict an appropriate sequence which leads to the simulation outcome of all tracks being skipped.

\subsection{Online Experiments}
We tested our model-based RL approach online at scale to assess the ability of the agent to power our real-world recommender systems. Our expectation is that the RL agent, trained offline against a non-sequential world model, should be able to produce satisfactory user-listening experiences assuming the world model accurately reflects user preferences. As the behavior of the agent is directly dependent on the accuracy of the user model, we also tested our agent online against the user model by itself (CWM-GMPC).

\paragraph{Experiment Settings.}
Similar to the experimental settings on the public dataset, we consider the case of a recommender system tasked with generating the best list of tracks for a user to listen to given a particular context (time of day, type of music requested, and so on).
Our recommender system has a similar objective in online experiments: to maximize user satisfaction. As a proxy for user satisfaction, we consider the completion count (and rate), \ie, the amount of tracks recommended by the policy that are completed by the user (and as a fraction on how many are started).
For this task we first train an agent offline using a non-sequential world model, CWM, and then deploy the agent online to serve recommendations to the users. In our production setting each user targeted by the test selects a playlist which is backed by a predefined track pool, and the system responds with an sub-selected ordered list of tracks of a size less than that of the pool.

The outcomes of our user model are primarily consumption-focused and summarize the probability of user-item interactions. Specifically, the CWM variant in this case has been optimized for three targets: \textit{completion}, \textit{skip} and \textit{listening duration}\footnote{To have a binary target, we discretize this value if the listening duration is longer than a predefined threshold $\tau$.}. The reward for the agent is computed as the sum of the probability of \textit{completion} for each track in an episode.
We compare our proposed agent against three alternative policies: 
\begin{enumerate}[label=\emph{(\roman*)}]
    \item \emph{Random}: a model to randomly sort the tracks with uniform distribution (\ie, all tracks have equal probability of appearing in a specific position); 
    \item \emph{Cosine Similarity}: a model which sorts the tracks based on the cosine similarity between predefined user and track embeddings;
    \item \emph{CWM-GMPC}: the user model ranking, which sorts the tracks based on the predicted probability of completion from the user. This is the same GMPC construction we use in the public dataset but the user-model targets have changed to match our real-world setting (as described above).
\end{enumerate}

All policies take as inputs features of the user that made the request and the pool-provided set of tracks with their associated features. The goal of each policy is to select and order a list of tracks from the pool that maximizes expected user satisfaction, which we measure by counting the number of tracks completed by the user. Note that the number of tracks that the user actually interacts with (\ie, that the user starts listening to) may be lower than the recommended number (\eg, if the user leaves the listening session early). For this reason, in what follows we measure not only the number of completed tracks, but also the rate of completion (the amount of tracks completed with respect to those that were actually started).

To study the effectiveness of our proposed approach we conducted a live A/B test on our production platform. We compared the previously described 4 policies to the default playlist generation model of our production platform by running a week long test. Users were randomly divided into five test cells, assigning the default model to \textit{control} group, and assigning the four policies to their respective \textit{treatment} groups. We collected a large sample of interaction data covering 2 million users, interacting with 2.8 million unique tracks across 4 million distinct sessions.

\paragraph{Offline-online correlation.}

\begin{figure}
    \centering
    \includegraphics[width=0.95\linewidth]{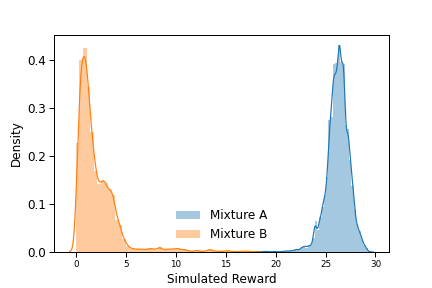}
    \caption{Reward distribution on simulated offline episodes as estimated by the SWM.}
    \label{fig:mixtures}
\end{figure}
\begin{figure}
    \centering
    \includegraphics[width=0.95\linewidth]{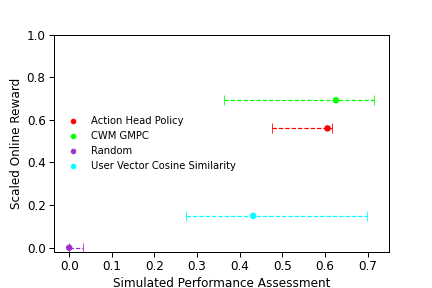}
    \caption{Simulated performance (estimated by the SWM) with respect to online reward (with true user responses) for the compared four different policies.}
    \label{fig:offline-online}
\end{figure}

First, we empirically assess the validity of our approach to simplify the action space of the agent through a world model which distills user behavior. 
\Cref{fig:offline-online} includes both offline performance estimates for each model alongside online results for the policies previously listed. Just like with the public dataset evaluation, we use an independent metric model in offline simulation to estimate policy performance, the sequential world model (SWM) described in \Cref{sec:public-results}.

For each policy we generate a list of recommended tracks for each evaluation episode. For each list we then compute the completion probabilities using the SWM and sum them. Finally, we average the total completion probabilities across the evaluation episodes to calculate the average reward for a policy.

Our simulator for real-world applications allows for a variety of settings including (but not limited to) the ability to simulate policy performance for different users and content. Different simulation settings and varying simulated episodes initializations can lead to difficult to summarize, multi-modal performance estimates from our simulator. 
For these results simply calculating the average reward of a policy across all episodes can fail to correctly summarize its performance because of skew, outliers, and the multi-modal distributions.
\Cref{fig:mixtures} highlights the bimodal distribution of the rewards for simulated episodes that can result from two common in user listening behaviors: lean-back (majority of tracks are completed) and active skipping (majority of tracks are skipped, or incomplete).
The offline performance metric is then computed using a normalized average of the modal returns from episode reward as follows. A Gaussian mixture model is used to separate the reward distributions, then values are logged and min-max scaled to $[0, 1]$ prior to each policy being assessed. A non-parametric bootstrap is used to obtain confidence intervals on our performance estimates for each policy. Such a procedure yields a more accurate reflection of a policy's simulated performance.

The offline performance expectations of the evaluated policies align with their online performance (\Cref{fig:offline-online}). Unsurprisingly, a random policy fails to perform well in both simulated and online settings. The \textit{Cosine Similarity} model has better performance than random, but it lacks the rich user and item features available to the other non-random policies. We see that the offline performance between the Action Head Policy (the agent) and CWM-GMPC is essentially the same. This is to some extent expected since the optimal policy for agent trained against the pointwise world model is, in fact, the greedy policy CWM-GMPC (\Cref{sec:world-model}).
Online results show a slight gap between these policies, but the difference is statistically indistinguishable. 

\paragraph{Online Evaluation.}

\begin{table}[]
    \centering
    \caption{Relative percent difference between world model (CWM) policy and control on online evaluation.}
    \label{tab:results_table}
    \begin{tabular}{cr}
\toprule
Metric (Per-Session Average): &               Relative \% Difference: \\
\midrule
             Completion-Count &   -2.9 (p: 0.88, CI: -39.38 | 33.59) \\
                    Total MSP &  -8.59 (p: 0.64, CI: -44.55 | 27.36) \\
                 MSP-Per-Item & -13.97 (p: 0.05, CI: -27.89 | -0.06) \\
                    Skip-Rate &   -5.49 (p: 0.42, CI: -18.77 | 7.79) \\
              Completion-Rate &    9.98 (p: 0.23, CI: -6.52 | 26.49) \\
Session Length (interactions) &  13.05 (p: 0.31, CI: -12.02 | 38.11) \\
\bottomrule
    \end{tabular}
\end{table}

\begin{table}[]
    \centering
    \caption{Relative percent difference between agent policy and control on online evaluation.}
    \label{table:agent_comp_tab}
    \begin{tabular}{cr}
\toprule
Metric (Per-Session Average): &              Relative \% Difference: \\
\midrule
             Completion-Count & 10.17 (p: 0.59, CI: -26.79 | 47.13) \\
                    Total MSP &  6.43 (p: 0.73, CI: -30.67 | 43.53) \\
                 MSP-Per-Item &  -5.62 (p: 0.46, CI: -20.69 | 9.44) \\
                    Skip-Rate &   -7.8 (p: 0.33, CI: -23.52 | 7.92) \\
              Completion-Rate &    5.39 (p: 0.6, CI: -14.8 | 25.58) \\
Session Length (interactions) &  8.87 (p: 0.53, CI: -19.02 | 36.75) \\
\bottomrule
    \end{tabular}
\end{table}

\Cref{tab:results_table,table:agent_comp_tab} include results on the CWM-GMPC and RL agent performance online, respectively. 
Online experiments that directly serve the user model (CWM-GMPC) demonstrate its performance is statistically indistinguishable from control for essentially all metrics of interest (\Cref{tab:results_table}). 
Note that of the metric comparisons in \Cref{tab:results_table}, most importantly, the probability of a user skipping or completing a track is statistically indistinguishable between the user model and control. This, along with the offline-online correlation analysis in \Cref{fig:offline-online}, shows how our approach to model user behavior is accurate in practice.

\begin{figure}[t]
    \centering
    \includegraphics[width=0.9\linewidth]{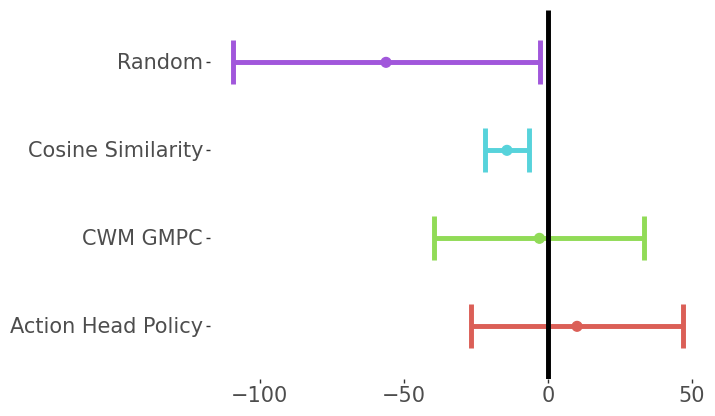}
    \caption{Percent difference in average completion count per-session relative to control.}
    \label{fig:comp_count}
\end{figure}
\begin{figure}[t]
    \centering
    \includegraphics[width=0.9\linewidth]{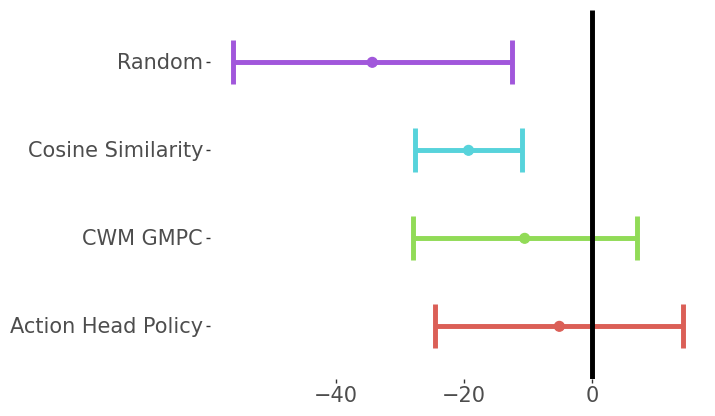}
    \caption{Percent difference in average completion rate per-session relative to control.}
    \label{fig:comp_rate}
\end{figure}

The performance of our trained agent and the additional comparative policies relative to control for completion count per session is shown in \Cref{fig:comp_count} and completion rate per session in \Cref{fig:comp_rate}.
Our expectation is that an agent trained against our world-model in an offline setting should at least mimic its online performance. Our results show that, online, our user model is statistically indistinguishable from control, so we expect to have similar results for our agent performance.
Hence, the objective of our online analysis is to demonstrate no statistically significant difference between control (playlist generator implemented in production) and the behavior of our agent. This is validated in our results, where both our user model ranking and the agent trained in simulation show results statistically indistinguishable from control, further validating our approach.

We also report the complete metric spread for the agent relative to control in \Cref{table:agent_comp_tab}. Note that although point estimates are different, the confidence intervals for the metric relative difference to control for the agent and user model ranking are more or less entirely overlapping for each metric. This highlights how the performance of CWM-GMPC and agent are statistically indistinguishable on real users, validating our approach to automatic playlist generation.

\paragraph{Offline Analysis to Predict Online Results.}

\begin{figure}
    \centering
    \includegraphics[width=0.95\linewidth]{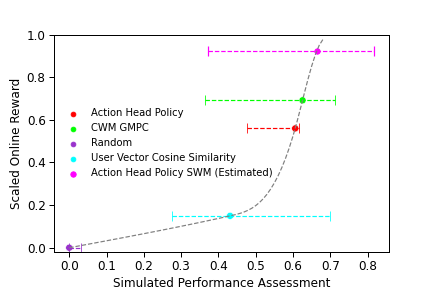}
    \caption{Online reward with respect to simulated performance with the addition of SWM online estimate.}
    \label{fig:offline-online-swm}
\end{figure}

One of the main goals of this work is to develop the ability to experiment with, train, and evaluate agent policies without exposing users to sub-optimal recommendations. We have seen how the agent trained against a pointwise user model (which we refer to as AH-CWM) is able to mimic its optimal policy online. However, the question of performance for the agent trained against more complex, sequential user models (SWM) with unknown optimal policies still remains. Training such agents brings additional complexity, as a sophisticated user model is required to be developed alongside an agent that can train against it effectively. In real scenarios, the implementation and online testing of a new policy is not straightforward, and needs to be backed up by some degree of assurance that it will not lead to a detrimental experience for the user to minimize the risk of users abandoning the platform.

For this reason, we consider our offline-online correlation analysis of \Cref{fig:offline-online} by adding the estimated offline performance assessment of an agent trained against the SWM (AH-SWM).
\Cref{fig:offline-online-swm} shows the offline performance of AH-SWM policy along with an estimate of its online performance derived via extrapolation. Such offline-online correlation analysis is fundamental in order to approximate what is the expected improvement that could be translated online. Based on these predictions we hypothesize an online improvement over both CWM-GMPC and AH-CWM policy performance. 
We argue this type of analysis of being relevant in practical applications of recommender system where online deployment requires sufficient expected improvement over existing baselines.

\section{Conclusion}\label{sec:conclusion}
In this paper we presented a reinforcement learning framework based on a simulated environment that we deployed in practice to efficiently use RL for playlist generation in a music streaming domain. We presented our use case which is different from standard slate recommendation task where usually the target is to select at maximum one item in the sequence. Here, instead, we assume to have a user-generated response for multiple items in the slate, making slate recommendation systems not directly applicable.
 
By making use of a learned world model that simulates user responses to the actions selected by the agent, we were able to train agents offline and evaluate their policies prior to exposure to real users. Online results show that even without further training
using online interactions the learned policy
does not result in loss of user satisfaction with respect to other baselines.


A further research direction is to improve the user behavior model for better agent training. 
The performance of the trained agent is influenced by the prediction accuracy of the user behavior model.
We can explore various ways to improve the user behavior model such as designing better user representations, exploring different neural network architectures such as transformers or increasing the robustness of prediction via techniques like dropout or ensemble methods.

\begin{acks}
We thank all the present and past team members involved in the projects that led to this publication for their hard work and dedication: 
Justin Carter, Elizabeth Kelly, Raj Kumar, Quan Yuan for their work on the backend and ML components and setting the infrastructure up for online tests; Jason Uh for guiding the project and coordinating online tests;
Dmitrii Moor, for his initial contribution to the world model concept; Mehdi Ben Ayed, for creating the infrastructure for RL experimentation and first offline and online tests; Marc Romeyn, Vinod Mohanan, Victor Delepine, Maya Hristakeva for laying the foundation of the work in its initial phase.
\end{acks}

\bibliographystyle{ACM-Reference-Format}
\balance
\bibliography{main-bib}

\appendix

\end{document}